%% file: iclr2020_conference.tex
\documentclass{article} % For LaTeX2e
\usepackage{iclr2020_conference,times}

\pdfoutput=1

\input{math_commands.tex}

\usepackage{graphicx}
\usepackage{hyperref}
\usepackage{subfigure}
\usepackage{url}
\usepackage{xcolor}

\title{Adaptive Prediction Timing for Electronic Health Records}

\author{Jacob Deasy\thanks{corresponding author JD, \url{https://github.com/jacobdeasy/dynamic-ehr}} \ \& Pietro Li\`{o}\\
Department of Computer Science and Technology\\
University of Cambridge\\
Cambridge, United Kingdom \\
\texttt{\{jd645,pl219\}@cam.ac.uk} \\
\And
Ari Ercole \\
Department of Medicine\\
University of Cambridge \\
Cambridge, United Kingdom \\
\texttt{ae105@cam.ac.uk} \\
}

\iclrfinalcopy % Uncomment for camera-ready version, but NOT for submission.
\begin{document}

\maketitle

\begin{abstract}
In realistic scenarios, multivariate timeseries evolve over case-by-case time-scales. This is particularly clear in medicine, where the rate of clinical events varies by ward, patient, and application. Increasingly complex models have been shown to effectively predict patient outcomes, but have failed to adapt granularity to these inherent temporal resolutions. As such, we introduce a novel, more realistic, approach to generating patient outcome predictions at an adaptive rate based on uncertainty accumulation in Bayesian recurrent models. We use a Recurrent Neural Network (RNN) and a Bayesian embedding layer with a new aggregation method to demonstrate adaptive prediction timing. Our model predicts more frequently when events are dense or the model is certain of event latent representations, and less frequently when readings are sparse or the model is uncertain. At 48 hours after patient admission, our model achieves equal performance compared to its static-windowed counterparts, while generating patient- and event-specific prediction timings that lead to improved predictive performance over the crucial first 12 hours of the patient stay.
\end{abstract}

\section{Introduction}
\label{introduction}

Over the past decade, machine learning, and deep learning in particular, have repeatedly demonstrated strong performance on a range of benchmark datasets (\cite{lecun2015deep,goodfellow2016deep}). These successes have since been transferred to medical data (\cite{esteva2019guide}), and specifically the domain of Electronic Health Records (EHRs, \cite{shickel2017deep}), where it is hoped research will lead to patient specification prognostication and diagnosis.

Recent state-of-the-art deep neural network (DNN) models for EHR data use recurrent models for sequence analysis which rely upon fixed prediction scheduling, carrying out extensive model analysis while overlooking the underlying choice of time window length (\cite{choi2016retain,rajkomar2018scalable,tomavsev2019clinically}). This early-stage modelling decision---necessitated by traditional Recurrent Neural Network (RNN, \cite{elman1990finding}) structure---loses patient information and timeseries granularity, and ignores the underlying timescales present in EHR data which have been shown to boost model performance (\cite{meiring2018optimal,che2018recurrent}).

In this paper, we introduce and analyse a novel method for adaptive prediction timing in the context of medical timeseries. Progress in variational Bayesian neural networks, facilitated by the reparameterisation trick (\cite{blundell2015weight,kucukelbir2017automatic}), has led to increased use of Bayesian embeddings of medical concepts (\cite{dusenberry2019analyzing}). We posit that embedding distribution uncertainty can be used to induce adaptive prediction timing. To this end, we explore the following:
\begin{enumerate}
    \item How can model uncertainty be related to prediction timing? (Section~\ref{sec:background}, paragraph 2)
    \item How does adaptive prediction timing affect model performance? (Table~\ref{tab:performance})
    \item How do prediction timings differ and adapt from fixed windows during training? (Figure~\ref{fig:timing_distribution})
    \item Does our model generalise to different clinical objectives and cohorts? (Table~\ref{tab:performance} and \ref{tab:performance_eICU})
\end{enumerate}

\paragraph{Contributions} We draw several significant conclusions from sequence modelling on the MIMIC-III (\cite{johnson2016mimic}) and the e-ICU (\cite{pollard2018eicu}, Section~\ref{sec:eicu}) datasets. We find that certainty rather than uncertainty, quantified by the precision of embedding distributions in a variational embedding layer, generates a natural measure of when to predict. In particular, we find that using the cumulative precision of embedding distributions encourages models to predict frequently when event sampling is dense and/or familiar to the model, and delays prediction when the model is uncertain about recent events---a more realistic approach to prediction timing. We demonstrate that the benefits of this model formulation do not impact negatively on final model performance. Finally, we highlight how adaptive prediction timing evolves over training to better utilise time periods of frequent events and produce correct predictions earlier in the patient stay.

\section{Background}
\label{sec:background}

\paragraph{Adaptive prediction timing} Recent, highly-cited, outcome prediction models for EHR timeseries (\cite{rajkomar2018scalable,tomavsev2019clinically}) have paid little attention to prediction timing, relegating window choice to the supplementary material, and omitting the question `\emph{When} is a good time to update model predictions?' from the line of enquiry. This approach contradicts evidence that modelling patient outcomes dynamically over time is beneficial (\cite{meiring2018optimal,deasy2019dynamic}), and overlooks literature on adaptive computation for RNNs (\cite{graves2016adaptive}). A few efforts to overcome irregular sampling have been made by learning interpolants and decay rates for individual variables across fixed time periods (\cite{che2018recurrent,shukla2019interpolation}), but these models still do not account for varying amounts of \emph{information} lost at the point of aggregation. The authors of \cite{liu2019learning} recently demonstrated that dynamic prediction, with a patient-specific temporal resolution found by classical min-max optimisation, outperforms the previous one-size-fits-all approach despite maintaining fixed windows. In this paper, we go further by arguing that the patient timeseries is, in fact, an \emph{information series} and it is more appropriate to evenly spread events based on model certainty. Our approach not only generates individualised prediction timing, but also \emph{event-specific} prediction timing---crucial in the highly heterogeneous and patient-specific environment of the hospital and the real world.

\paragraph{Embedding precision} In a Bayesian RNN, the variational inference approach to learning the weights $\bm{w}$ of the approximate model $q(\bm{w}|\bm{\theta})$, dictates the use of factorised weight posteriors $q(\bm{w}|\bm{\theta})=\prod_{i}q(\bm{w}_{i}|\bm{\theta}_{i})$. When the $\bm{w}_{i}$ follow a multivariate Gaussian distribution, with learnable mean vector $\bm{\mu}$ and diagonal covariance matrix $\bm{\Sigma}$, we have
\begin{align}
    \bm{w}\sim\mathcal{N}(\bm{\mu}, \bm{\Sigma}) \implies
    \bm{w}\sim\mathcal{N}(\bm{\mu}, \bm{\Phi}^{-1}),
\end{align}
where we drop the index for ease of notation and note that $\bm{\Phi}$, the inverse of the covariance matrix $\bm{\Sigma}$, is the precision matrix of the multivariate normal. The precision of each embedding distribution is, therefore, defined by
\begin{align}
    \textrm{Precision}(\bm{w}) = \textrm{det}(\bm{\Phi}) = \prod\limits_{j}\bm{\Sigma}_{jj}^{-2},
\end{align}
and is as a measure of model certainty (\cite{degroot2005optimal}).

\section{Adaptive prediction timing}
\label{medical_uncertainty}

\paragraph{Clinical objectives} We analyse the performance of a novel recurrent model on in-hospital mortality and long length of stay (defined as greater than 7 days, \cite{rajkomar2018scalable}) prediction, both at 48 hours after admission. Dynamic mortality risk estimation helps summarise the patient state, predict patient trajectory, and is the subject of multiple clinical severity scores (\cite{rapsang2014scoring}). Equally, long length of stay estimation enables ward management planning and resource allocation across the hospital. We employ the MIMIC-III database (\cite{johnson2016mimic}), an EHR dataset collected from 46,520 patients admitted to intensive care units (ICUs) at Beth Israel Deaconess Medical Center. We embed all chart, lab, and output events as described in \cite{deasy2019dynamic}, utilise our adaptive prediction timing aggregation step, feed the output to a layer-normalised LSTM (\cite{ba2016layer,hochreiter1997long}), and perform an affine transformation before a sigmoid output activation.

\paragraph{Method} For a given patient sequence, comprised of time points and events $\{t_{i}, x_{i}\}_{0\leq i \leq n}$, instead of aggregating events into fixed time intervals, we first sample the corresponding sequence of event embedding distributions to obtain the sequence $\{\bm{w}_{i}\}_{0\leq i \leq n}$. To separate these samples into intervals, we then use the embedding distributions to generate a corresponding cumulative precision sequence
\begin{align}
    \bm{p}^{*}_{k} = \sum\limits_{i=0}^{k}\bm{p_{i}} = \sum\limits_{i=0}^{k}\left(\prod\limits_{j}\bm{(\Sigma_i)}_{jj}^{-2}\right),
\end{align}
and separate this sequence into \emph{equi-precise} aggregation windows which evolve as embedding distributions are refined by training. At no point does our model make use of event timestamps. We implement this in a vectorised manner to handle batches of size greater than one.

Our models were trained by minimising the Kullback-Leibler (KL) divergence between the approximate posterior and the actual, intractable, posterior via the reparameterisation trick. Equivalent to minimising an expectation over the negative log-likelihood term plus a KL regularisation term
\begin{align}
    \mathcal{L}(\bm{\theta}) &= \textrm{KL}[q(\bm{w}|\bm{\theta})\parallel p(\bm{w}|\bm{y},\bm{X})]\\
    &\propto \textrm{KL}[q(\bm{w}|\bm{\theta})\parallel p(\bm{w})] - \mathbb{E}_{q}[\ln p(\bm{y}|\bm{X},\bm{w})].
\end{align}

\section{Experiments}
\label{others}

\paragraph{Clinical tasks}
For our clinical tasks, to assess predictive performance, we measure area under the precision-recall curve (AUPRC), area under the receiver operating characteristic curve (AUROC), and, as there is a strong class imbalance (see Table~\ref{tab:dataset_information}), Matthews Correlation Coefficient (MCC). Table~\ref{tab:performance} shows the mean and standard deviation of the metrics at 48 hours after admission. Throughout, we re-sample the embedding layer of the variational models 100 times and bootstrapped ensembles of 10 deterministic models with 1000 re-samples to generate error measurements. Despite the underlying change in aggregation mechanism, the final performance of our model is very strong, inline with the literature (\cite{dusenberry2019analyzing}), and the predictions are well-calibrated (see Figure~\ref{fig:calibration}), demonstrating strong performance on both clinical tasks. We also verify model generalisation for both tasks on the eICU dataset (\cite{pollard2018eicu}) in Table~\ref{tab:performance_eICU}.

\begin{table}[t]
\caption{Mean (and standard deviation) of metrics for the adaptive prediction timing model on the binary mortality and long length of stay tasks---max MCC over 100 thresholds is reported. Our model displays strong predictive performance, with robust generalisation to the held-out test set.}
\label{tab:performance}
\begin{center}
\begin{tabular}{llll}
\multicolumn{1}{c}{\bf TASK}  &\multicolumn{1}{c}{\bf METRIC}  &\multicolumn{1}{c}{\bf VALIDATION} &\multicolumn{1}{c}{\bf TEST}
\\ \hline \\
Mortality           & AUPRC & 0.576 ($\pm$0.013) & 0.556 ($\pm$0.011)\\
                    & AUROC & 0.897 ($\pm$0.003) & 0.879 ($\pm$0.004)\\
                    & MCC   & 0.510 ($\pm$0.011) & 0.496 ($\pm$0.012)\\
Long length of stay & AUPRC & 0.614 ($\pm$0.008) & 0.566 ($\pm$0.009)\\
                    & AUROC & 0.834 ($\pm$0.004) & 0.830 ($\pm$0.004)\\
                    & MCC   & 0.494 ($\pm$0.009) & 0.465 ($\pm$0.009)
\end{tabular}
\end{center}
\end{table}

\paragraph{Model variants}
We assess our model against a range of differing models in Table~\ref{tab:model_variants}. We name these models based on whether their embedding layer is deterministic or Bayesian, and whether their prediction timings are based on fixed timing (no prefix), fixed event count (\#), or fixed cumulative precision ($p^{*}$). We include a model which aggregates by event count, to isolate the effect of our model's adaptation, and conclude it has a marginally negative effect on generalisation, which must be weighed against the advantages of improved temporal resolution during eventful periods.

\begin{table}[t]
\caption{Performance comparison between different model variants for the binary mortality prediction tasks on the MIMIC-III dataset.}
\label{tab:model_variants}
\begin{center}
\begin{tabular}{lllll}
\multicolumn{1}{c}{\bf MODEL}  &\multicolumn{1}{c}{\bf VAL. AUPRC}  &\multicolumn{1}{c}{\bf VAL AUROC} &\multicolumn{1}{c}{\bf TEST AUPRC} &\multicolumn{1}{c}{\bf TEST AUROC}
\\ \hline \\
Deterministic LSTM   & 0.592 ($\pm$0.010) & 0.887  ($\pm$0.008) & 0.595  ($\pm$0.012) & 0.889  ($\pm$0.006) \\
Deterministic \#-LSTM & 0.602 ($\pm$0.012) & 0.883 ($\pm$0.006) & 0.578 ($\pm$0.011) &  0.883 ($\pm$0.005)\\
Bayesian LSTM         & 0.574 ($\pm$0.013) & 0.886 ($\pm$0.004) & 0.571 ($\pm$0.012) & 0.883 ($\pm$0.004)\\
Bayesian \#-LSTM      & 0.582 ($\pm$0.011) & 0.889 ($\pm$0.004) & 0.573 ($\pm$0.013) & 0.881 ($\pm$0.004)\\
Bayesian $p^{*}$-LSTM & 0.576 ($\pm$0.013) & 0.897 ($\pm$0.003) & 0.556 ($\pm$0.011) & 0.879 ($\pm$0.004)
\end{tabular}
\end{center}
\end{table}

\begin{figure}
    \centering
    \subfigure{\label{fig:timing_distribution_left}\includegraphics[width=0.5\textwidth]{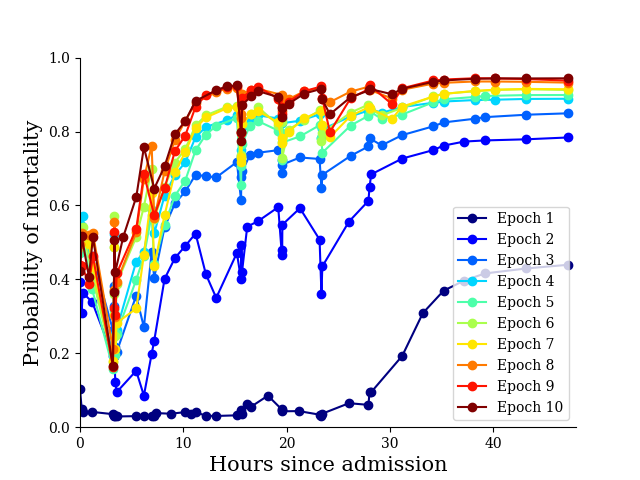}}%
    \subfigure{\label{fig:timing_distribution_right}\includegraphics[width=0.5\textwidth]{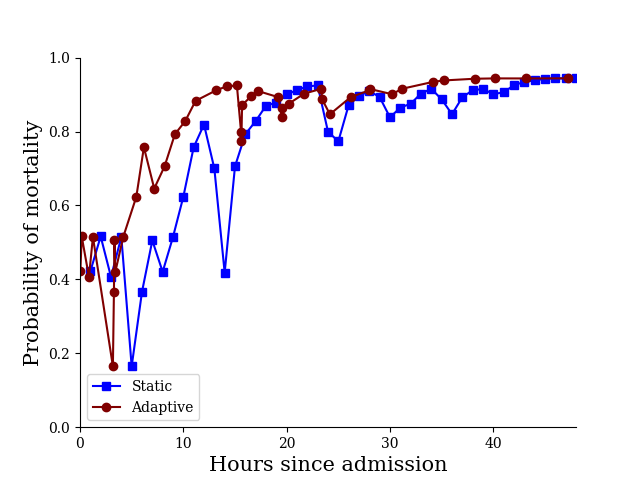}}
    \caption{\textbf{Left}--Mortality probability evolving through training for a patient that went on to die. \textbf{Right}--The adaptive timing model predicts mortality earlier than the static window model.}
    \label{fig:timing_distribution}
\end{figure}

\paragraph{Early predictive power} In Figure~\ref{fig:timing_distribution}, we compare mortality risk prediction of the static model with our adaptive model for a patient who went on to die in hospital. The more fine-grained prediction timings learnt by the variational model, displayed in Figure~\ref{fig:timing_distribution}-left, led to earlier prediction of mortality compared to the static model in Figure~\ref{fig:timing_distribution}-right due to patient-specific segmentation of the timeseries. As most patients in the ICU have many additional readings taken in the first hours of their stay (e.g. admission information and medical history), which clinicians use to more frequently update their opinion of the patient state, this is a more realistic approach to outcome prediction.

\begin{figure}
    \centering
    \subfigure{\label{fig:pmort_training2}\includegraphics[width=0.5\textwidth]{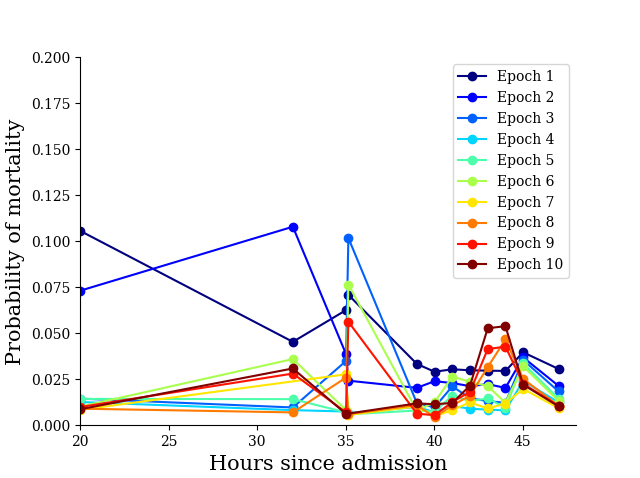}}%
    \subfigure{\label{fig:time_density3}\includegraphics[width=0.5\textwidth]{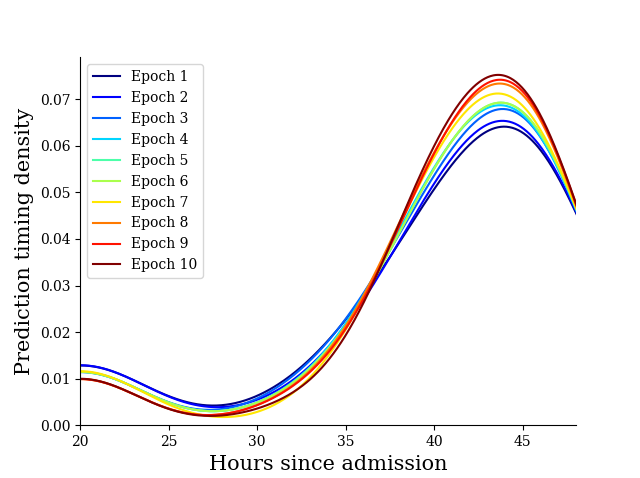}}
    \caption{Example of how the timing of model predictions coalesce through training for a patient with frequent event recordings towards the end of their stay.}
    \label{fig:prediction_timing}
\end{figure}

\paragraph{Prediction timing evolution} In Figure~\ref{fig:prediction_timing}, we demonstrate the evolution of prediction timing for a different patient. In Figure~\ref{fig:prediction_timing}-right, the prediction timing distribution can be seen to focus on a particularly event-dense period for this patient as it learns to be more certain about the embedding distributions of particular clinical events. This suggests our model would adapt well to more extreme shifts in granularity such as stays which include either surgical or emergency interventions.

\newpage

\bibliography{iclr2020_conference}
\bibliographystyle{iclr2020_conference}

\newpage

\appendix
\section{Appendix}
\renewcommand{\thefigure}{\thesection.\arabic{figure}}
\setcounter{figure}{0}
\renewcommand{\thetable}{\thesection.\arabic{table}}
\setcounter{table}{0}

\subsection{Additional dataset information}

After applying the pipeline described in \cite{deasy2019dynamic}, our subset of the MIMIC-III dataset contained 21,143 patient stays including the following demographic and outcome ratios.
\begin{table}[ht]
\caption{Dataset information.}
\label{tab:dataset_information}
\begin{center}
\begin{tabular}{llll}
\multicolumn{1}{c}{\bf } &\multicolumn{1}{c}{\bf TRAIN (\%)} &\multicolumn{1}{c}{\bf VALIDATION (\%)} &\multicolumn{1}{c}{\bf TEST (\%)}
\\ \hline \\
Male                  & $9329/16913$ (55.2\%) & $1140/2115$ (53.9\%) & $1162/2115$ (54.9\%)\\
In-hospital mortality & $2237/16913$ (13.2\%) & $280/2115$  (13.2\%) & $280/2115$  (13.2\%)\\
Long length of stay   & $3698/16913$ (21.9\%) & $512/2115$  (24.2\%) & $462/2115$  (21.8\%)
\end{tabular}
\end{center}
\end{table}

\subsection{eICU Collaborative Research Database}
\label{sec:eicu}
As a proof of concept that our findings generalise, we also experiment with a small subset of the aperiodic vital sign readings in the eICU Collaborative Research Database (eICU) dataset (\cite{pollard2018eicu}), another publicly available EHR dataset. Results are displayed in Table~\ref{tab:performance_eICU}.
\begin{table}[ht]
\caption{Performance of the adaptive prediction timing model for the binary mortality and long length of stay tasks on the eICU dataset.}
\label{tab:performance_eICU}
\begin{center}
\begin{tabular}{llll}
\multicolumn{1}{c}{\bf TASK}  &\multicolumn{1}{c}{\bf METRIC}  &\multicolumn{1}{c}{\bf VALIDATION} &\multicolumn{1}{c}{\bf TEST}
\\ \hline \\
Mortality           & AUPRC & 0.253 ($\pm$ 0.012) & 0.244 ($\pm$ 0.010)\\
                    & AUROC & 0.705 ($\pm$ 0.005) & 0.701 ($\pm$ 0.006)\\
Long length of stay & AUPRC & 0.217 ($\pm$ 0.010) & 0.205 ($\pm$ 0.012)\\
                    & AUROC & 0.610 ($\pm$ 0.005) & 0.600 ($\pm$ 0.007)
\end{tabular}
\end{center}
\end{table}

\subsection{Model calibration}
\begin{figure}[ht]
    \centering
    \subfigure[Deterministic model.]{\label{fig:calibration_deterministic}\includegraphics[width=0.5\textwidth]{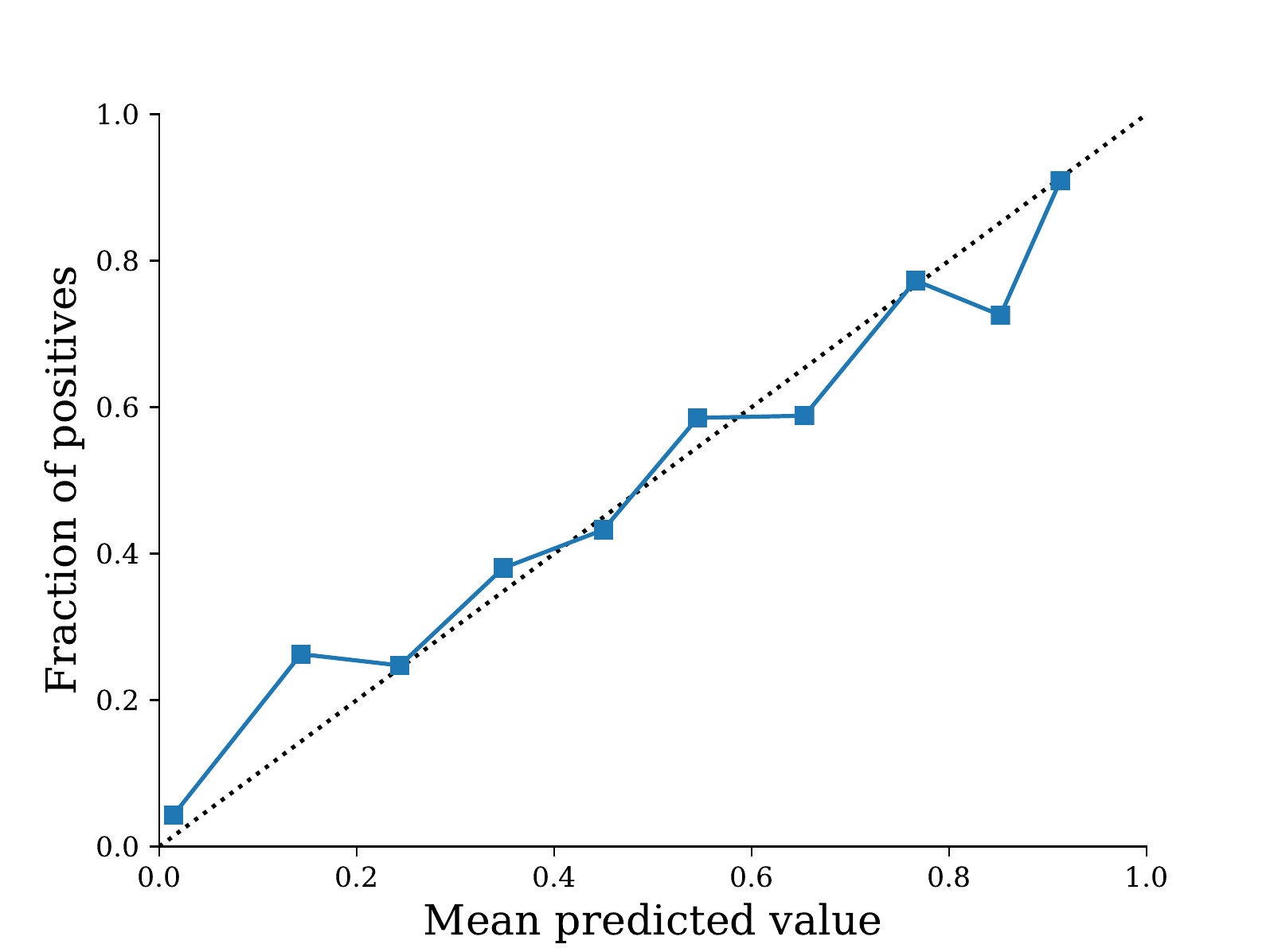}}%
    \subfigure[Variational model.]{\label{fig:calibration_variational}\includegraphics[width=0.5\textwidth]{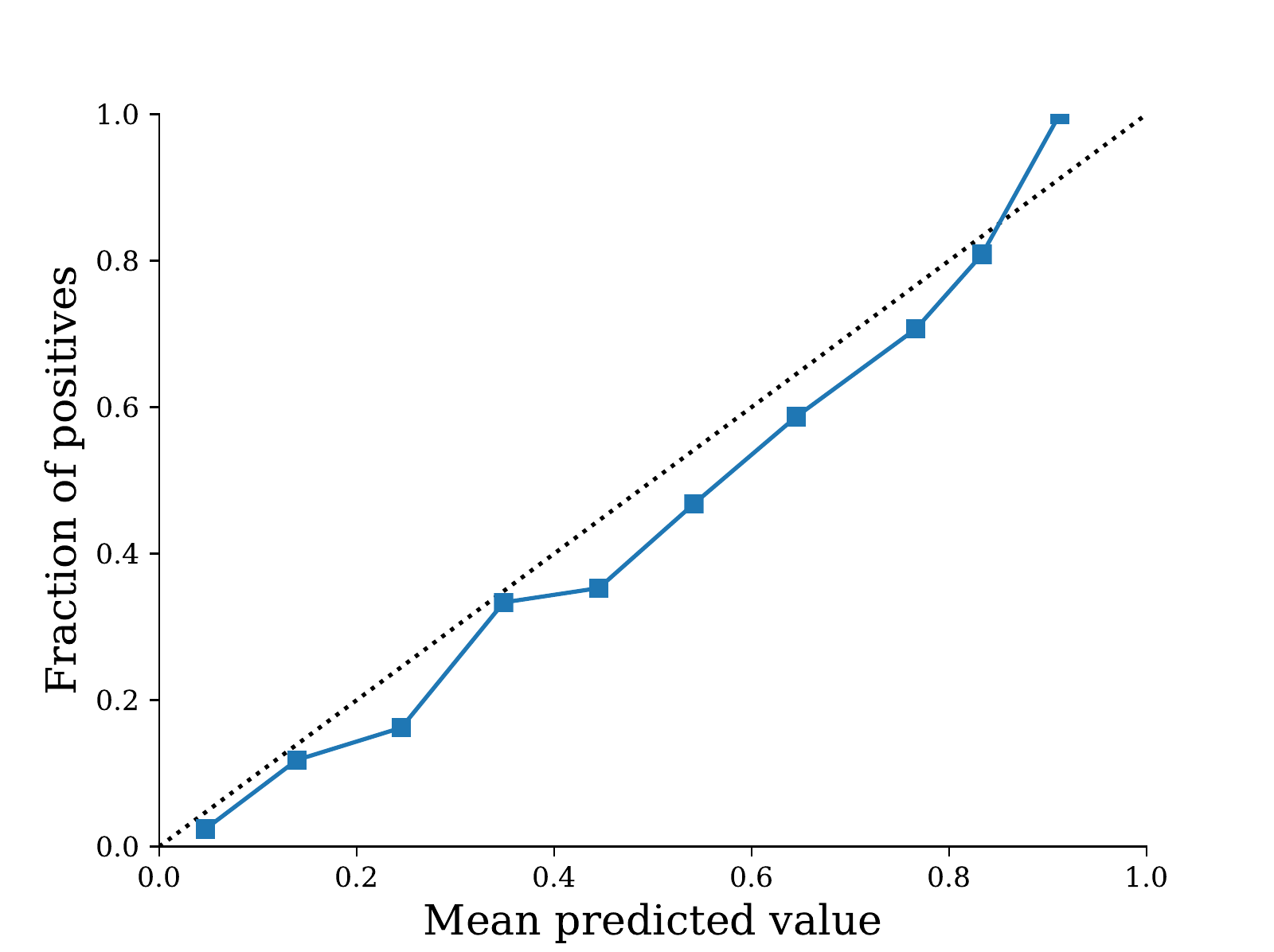}}
    \caption{Model calibration curves.}
    \label{fig:calibration}
\end{figure}

\subsection{Additional training details}
As in \cite{deasy2019dynamic}, our model uses embeddings of raw, unprocessed medical concepts in order to avoid any loss of extraneous information. Continuous variables were binned into 10 discrete categories assigned by quantile, and discrete variables were left untouched. Missing events were also embedded as discrete categories so our model makes use of \emph{informative missingness} (\cite{che2018recurrent}). Following on from the embedding layer, we noted during experimentation that a layer-normalised LSTM variant led to a considerable increase in performance. We initialise our LSTM with glorot initialisation (\cite{glorot2010understanding}) for the input-to-hidden matrices, orthogonal initialisation for the hidden-to-hidden matrices, and set the forget gate bias to 1 before training (\cite{jozefowicz2015empirical}).

For each model, we search over the space of hyperparameters defined in Table~\ref{tab:hyperparameter_options} using the Python package \texttt{wandb} from Weights and Biases. The search was performed using Bayesian optimisation to minimise validation set loss over both discrete and continuous variables. We use HyperBand early stopping (\cite{li2017hyperband}), with $s=2$, $\eta=3$, and $max\_iter=10$ to expedite optimisation. We fix batch size at 64 for all models.

Models were implemented in PyTorch 1.4, and trained on a Nvidia Titan X using Adam optimisation (\cite{kingma2014adam}). Both e-ICU and MIMIC-III datasets were split into train, validation, and test sets in a ratio of 8:1:1.

\begin{table}[ht]
\caption{Hyperparameter ranges and options.}
\label{tab:hyperparameter_options}
\begin{center}
\begin{tabular}{ll}
\multicolumn{1}{c}{\bf HYPERPARAMETER}  &\multicolumn{1}{c}{\bf RANGE}
\\ \hline \\
Learning rate       & [0.00001, 0.1]\\
$L_{2}$ regularisation coefficient & [0.0, 0.01]\\
Prior standard deviation (Bayesian only) & [0.1, 1.0]\\
Embedding dimension & [16, 32, 48, 64]\\
LSTM hidden dimension & [16, 32, 64, 128, 256, 512]
\end{tabular}
\end{center}
\end{table}

\end{document}

%% file: math_commands.tex
\usepackage{amsmath,amsfonts,bm}

\def\eqref#1{equation~\ref{#1}}
\def\1{\bm{1}}

\DeclareMathAlphabet{\mathsfit}{\encodingdefault}{\sfdefault}{m}{sl}
\SetMathAlphabet{\mathsfit}{bold}{\encodingdefault}{\sfdefault}{bx}{n}

%% file: iclr2020_conference.bbl
\begin{thebibliography}{27}
\providecommand{\natexlab}[1]{#1}
\providecommand{\url}[1]{\texttt{#1}}
\expandafter\ifx\csname urlstyle\endcsname\relax
  \providecommand{\doi}[1]{doi: #1}\else
  \providecommand{\doi}{doi: \begingroup \urlstyle{rm}\Url}\fi

\bibitem[Ba et~al.(2016)Ba, Kiros, and Hinton]{ba2016layer}
Jimmy~Lei Ba, Jamie~Ryan Kiros, and Geoffrey~E Hinton.
\newblock Layer normalization.
\newblock \emph{arXiv preprint arXiv:1607.06450}, 2016.

\bibitem[Blundell et~al.(2015)Blundell, Cornebise, Kavukcuoglu, and
  Wierstra]{blundell2015weight}
Charles Blundell, Julien Cornebise, Koray Kavukcuoglu, and Daan Wierstra.
\newblock Weight uncertainty in neural networks.
\newblock \emph{arXiv preprint arXiv:1505.05424}, 2015.

\bibitem[Che et~al.(2018)Che, Purushotham, Cho, Sontag, and
  Liu]{che2018recurrent}
Zhengping Che, Sanjay Purushotham, Kyunghyun Cho, David Sontag, and Yan Liu.
\newblock Recurrent neural networks for multivariate time series with missing
  values.
\newblock \emph{Scientific reports}, 8\penalty0 (1):\penalty0 6085, 2018.

\bibitem[Choi et~al.(2016)Choi, Bahadori, Sun, Kulas, Schuetz, and
  Stewart]{choi2016retain}
Edward Choi, Mohammad~Taha Bahadori, Jimeng Sun, Joshua Kulas, Andy Schuetz,
  and Walter Stewart.
\newblock Retain: An interpretable predictive model for healthcare using
  reverse time attention mechanism.
\newblock In \emph{Advances in Neural Information Processing Systems}, pp.\
  3504--3512, 2016.

\bibitem[Deasy et~al.(2019)Deasy, Li{\`o}, and Ercole]{deasy2019dynamic}
Jacob Deasy, Pietro Li{\`o}, and Ari Ercole.
\newblock Dynamic survival prediction in intensive care units from
  heterogeneous time series without the need for variable selection or
  pre-processing.
\newblock \emph{arXiv preprint arXiv:1909.07214}, 2019.

\bibitem[DeGroot(2005)]{degroot2005optimal}
Morris~H DeGroot.
\newblock \emph{Optimal statistical decisions}, volume~82.
\newblock John Wiley \& Sons, 2005.

\bibitem[Dusenberry et~al.(2019)Dusenberry, Tran, Choi, Kemp, Nixon, Jerfel,
  Heller, and Dai]{dusenberry2019analyzing}
Michael~W Dusenberry, Dustin Tran, Edward Choi, Jonas Kemp, Jeremy Nixon,
  Ghassen Jerfel, Katherine Heller, and Andrew~M Dai.
\newblock Analyzing the role of model uncertainty for electronic health
  records.
\newblock \emph{arXiv preprint arXiv:1906.03842}, 2019.

\bibitem[Elman(1990)]{elman1990finding}
Jeffrey~L Elman.
\newblock Finding structure in time.
\newblock \emph{Cognitive science}, 14\penalty0 (2):\penalty0 179--211, 1990.

\bibitem[Esteva et~al.(2019)Esteva, Robicquet, Ramsundar, Kuleshov, DePristo,
  Chou, Cui, Corrado, Thrun, and Dean]{esteva2019guide}
Andre Esteva, Alexandre Robicquet, Bharath Ramsundar, Volodymyr Kuleshov, Mark
  DePristo, Katherine Chou, Claire Cui, Greg Corrado, Sebastian Thrun, and Jeff
  Dean.
\newblock A guide to deep learning in healthcare.
\newblock \emph{Nature medicine}, 25\penalty0 (1):\penalty0 24, 2019.

\bibitem[Glorot \& Bengio(2010)Glorot and Bengio]{glorot2010understanding}
Xavier Glorot and Yoshua Bengio.
\newblock Understanding the difficulty of training deep feedforward neural
  networks.
\newblock In \emph{Proceedings of the thirteenth international conference on
  artificial intelligence and statistics}, pp.\  249--256, 2010.

\bibitem[Goodfellow et~al.(2016)Goodfellow, Bengio, Courville, and
  Bengio]{goodfellow2016deep}
Ian Goodfellow, Yoshua Bengio, Aaron Courville, and Yoshua Bengio.
\newblock \emph{Deep learning}, volume~1.
\newblock MIT Press, 2016.

\bibitem[Graves(2016)]{graves2016adaptive}
Alex Graves.
\newblock Adaptive computation time for recurrent neural networks.
\newblock \emph{arXiv preprint arXiv:1603.08983}, 2016.

\bibitem[Hochreiter \& Schmidhuber(1997)Hochreiter and
  Schmidhuber]{hochreiter1997long}
Sepp Hochreiter and J{\"u}rgen Schmidhuber.
\newblock Long short-term memory.
\newblock \emph{Neural computation}, 9\penalty0 (8):\penalty0 1735--1780, 1997.

\bibitem[Johnson et~al.(2016)Johnson, Pollard, Shen, Li-wei, Feng, Ghassemi,
  Moody, Szolovits, Celi, and Mark]{johnson2016mimic}
Alistair~EW Johnson, Tom~J Pollard, Lu~Shen, H~Lehman Li-wei, Mengling Feng,
  Mohammad Ghassemi, Benjamin Moody, Peter Szolovits, Leo~Anthony Celi, and
  Roger~G Mark.
\newblock Mimic-iii, a freely accessible critical care database.
\newblock \emph{Scientific data}, 3:\penalty0 160035, 2016.

\bibitem[Jozefowicz et~al.(2015)Jozefowicz, Zaremba, and
  Sutskever]{jozefowicz2015empirical}
Rafal Jozefowicz, Wojciech Zaremba, and Ilya Sutskever.
\newblock An empirical exploration of recurrent network architectures.
\newblock In \emph{International conference on machine learning}, pp.\
  2342--2350, 2015.

\bibitem[Kingma \& Ba(2014)Kingma and Ba]{kingma2014adam}
Diederik~P Kingma and Jimmy Ba.
\newblock Adam: A method for stochastic optimization.
\newblock \emph{arXiv preprint arXiv:1412.6980}, 2014.

\bibitem[Kucukelbir et~al.(2017)Kucukelbir, Tran, Ranganath, Gelman, and
  Blei]{kucukelbir2017automatic}
Alp Kucukelbir, Dustin Tran, Rajesh Ranganath, Andrew Gelman, and David~M Blei.
\newblock Automatic differentiation variational inference.
\newblock \emph{The Journal of Machine Learning Research}, 18\penalty0
  (1):\penalty0 430--474, 2017.

\bibitem[LeCun et~al.(2015)LeCun, Bengio, and Hinton]{lecun2015deep}
Yann LeCun, Yoshua Bengio, and Geoffrey Hinton.
\newblock Deep learning.
\newblock \emph{nature}, 521\penalty0 (7553):\penalty0 436, 2015.

\bibitem[Li et~al.(2017)Li, Jamieson, DeSalvo, Rostamizadeh, and
  Talwalkar]{li2017hyperband}
Lisha Li, Kevin Jamieson, Giulia DeSalvo, Afshin Rostamizadeh, and Ameet
  Talwalkar.
\newblock Hyperband: A novel bandit-based approach to hyperparameter
  optimization.
\newblock \emph{The Journal of Machine Learning Research}, 18\penalty0
  (1):\penalty0 6765--6816, 2017.

\bibitem[Liu et~al.(2019)Liu, Li, Hu, Shi, Wang, Tang, and
  Zhang]{liu2019learning}
Luchen Liu, Haoran Li, Zhiting Hu, Haoran Shi, Zichang Wang, Jian Tang, and
  Ming Zhang.
\newblock Learning hierarchical representations of electronic health records
  for clinical outcome prediction.
\newblock \emph{arXiv preprint arXiv:1903.08652}, 2019.

\bibitem[Meiring et~al.(2018)Meiring, Dixit, Harris, MacCallum, Brealey,
  Watkinson, Jones, Ashworth, Beale, Brett, et~al.]{meiring2018optimal}
Christopher Meiring, Abhishek Dixit, Steve Harris, Niall~S MacCallum, David~A
  Brealey, Peter~J Watkinson, Andrew Jones, Simon Ashworth, Richard Beale,
  Stephen~J Brett, et~al.
\newblock Optimal intensive care outcome prediction over time using machine
  learning.
\newblock \emph{PloS one}, 13\penalty0 (11):\penalty0 e0206862, 2018.

\bibitem[Pollard et~al.(2018)Pollard, Johnson, Raffa, Celi, Mark, and
  Badawi]{pollard2018eicu}
Tom~J Pollard, Alistair~EW Johnson, Jesse~D Raffa, Leo~A Celi, Roger~G Mark,
  and Omar Badawi.
\newblock The eicu collaborative research database, a freely available
  multi-center database for critical care research.
\newblock \emph{Scientific data}, 5, 2018.

\bibitem[Rajkomar et~al.(2018)Rajkomar, Oren, Chen, Dai, Hajaj, Hardt, Liu,
  Liu, Marcus, Sun, et~al.]{rajkomar2018scalable}
Alvin Rajkomar, Eyal Oren, Kai Chen, Andrew~M Dai, Nissan Hajaj, Michaela
  Hardt, Peter~J Liu, Xiaobing Liu, Jake Marcus, Mimi Sun, et~al.
\newblock Scalable and accurate deep learning with electronic health records.
\newblock \emph{NPJ Digital Medicine}, 1\penalty0 (1):\penalty0 18, 2018.

\bibitem[Rapsang \& Shyam(2014)Rapsang and Shyam]{rapsang2014scoring}
Amy~Grace Rapsang and Devajit~C Shyam.
\newblock Scoring systems in the intensive care unit: a compendium.
\newblock \emph{Indian journal of critical care medicine: peer-reviewed,
  official publication of Indian Society of Critical Care Medicine},
  18\penalty0 (4):\penalty0 220, 2014.

\bibitem[Shickel et~al.(2017)Shickel, Tighe, Bihorac, and
  Rashidi]{shickel2017deep}
Benjamin Shickel, Patrick~James Tighe, Azra Bihorac, and Parisa Rashidi.
\newblock Deep ehr: a survey of recent advances in deep learning techniques for
  electronic health record (ehr) analysis.
\newblock \emph{IEEE journal of biomedical and health informatics}, 22\penalty0
  (5):\penalty0 1589--1604, 2017.

\bibitem[Shukla \& Marlin(2019)Shukla and Marlin]{shukla2019interpolation}
Satya~Narayan Shukla and Benjamin~M Marlin.
\newblock Interpolation-prediction networks for irregularly sampled time
  series.
\newblock \emph{arXiv preprint arXiv:1909.07782}, 2019.

\bibitem[Toma{\v{s}}ev et~al.(2019)Toma{\v{s}}ev, Glorot, Rae, Zielinski,
  Askham, Saraiva, Mottram, Meyer, Ravuri, Protsyuk,
  et~al.]{tomavsev2019clinically}
Nenad Toma{\v{s}}ev, Xavier Glorot, Jack~W Rae, Michal Zielinski, Harry Askham,
  Andre Saraiva, Anne Mottram, Clemens Meyer, Suman Ravuri, Ivan Protsyuk,
  et~al.
\newblock A clinically applicable approach to continuous prediction of future
  acute kidney injury.
\newblock \emph{Nature}, 572\penalty0 (7767):\penalty0 116--119, 2019.

\end{thebibliography}
